\documentclass{article}

\usepackage[preprint]{neurips_2020}

\usepackage[utf8]{inputenc} %
\usepackage[T1]{fontenc}    %
\usepackage[hidelinks]{hyperref}       %
\usepackage{url}            %
\usepackage{booktabs}       %
\usepackage{amsfonts}       %
\usepackage{nicefrac}       %
\usepackage{microtype}      %
\usepackage{caption}
\usepackage{subcaption}
\usepackage{amsmath}
\usepackage{tikz}
\usetikzlibrary{calc,positioning,shapes,automata}
\usepackage{enumitem}

\title{Learning to Reason in Large Theories \\ without Imitation}

\author{%
  Kshitij Bansal \\
  Google Research \\
  \texttt{kbk@google.com}
  \And
  Christian Szegedy \\
  Google Research \\
  \texttt{szegedy@google.com}
  \And
  Markus N.~Rabe \\
  Google Research \\
  \texttt{mrabe@google.com}
  \AND
  Sarah M.~Loos \\
  Google Research \\
  \texttt{smloos@google.com}
  \And
  Viktor Toman \\
  IST Austria \\
  \texttt{vtoman@ist.ac.at}
}

\newcommand{\Nats}{\mathbb{N}}

\newcommand{\myblue}{blue!85!black}
\newcommand{\myred}{red!85!black}

\begin{document}

\maketitle

\begin{abstract}
In this paper, we demonstrate how to do automated theorem proving in the presence of a large knowledge base of potential premises without learning from human proofs.
We suggest an exploration mechanism that mixes in additional premises selected by a tf-idf (term frequency-inverse document frequency) based lookup in a deep reinforcement learning scenario.
This helps with exploring and learning which premises are relevant for proving a new theorem.
Our experiments show that the theorem prover trained with this exploration mechanism outperforms provers that are trained only on human proofs.
It approaches the performance of a prover trained by a combination of imitation and reinforcement learning.
We perform multiple experiments to understand the importance of the underlying assumptions that make our exploration approach work, thus explaining our design choices.
\end{abstract}

\section{Introduction}

Theorem proving is a challenging benchmark for automated reasoning, and is an important milestone on the road to demonstrating that machine learning can produce a deep understanding of abstract concepts. 
In the long run, automated mathematical reasoning may become an important tool in engineering and scientific discovery.
Due to their success in many other areas, neural networks have recently been considered as a way to guide theorem proving~\citep{alemi2016deepmath,gauthier2017tactictoe,loos2017deep,huang2018gamepad,bansal2019holist,paliwal2020graph} and demonstrate approximate mathematical reasoning abilities in latent space~\citep{lee2020mathematical}.

While there is only a relatively small number of fundamental proof rules (or proof tactics) applicable at any point in a proof, there is a very large number of premises (i.e., previously proven theorems and lemmas) that could be invoked.
The main problem of \emph{reasoning in large theories} thus is to identify the premises relevant in the current context and thereby reduce the branching factor of the proof search to a manageable size.
This problem will become even more pronounced over time, as the theorem provers become more powerful, growing the number of available premises.

Previous works have relied on human proofs to either directly provide
or learn~\citep{bansal2019holist,paliwal2020graph} which premises are relevant to the current proof.
However, any open-ended system for mathematical reasoning needs to be able to learn which premises are relevant without human guidance.
In this work, we thus consider the problem of training a theorem prover \emph{without access to human proofs}.

We demonstrate that training the theorem prover without human data succeeds when we use reinforcement learning and help exploration by selecting a portion of the premises with a \mbox{tf-idf~\citep{manning2008introduction}} metric.
The theorem prover thereby learns to prove more theorems than the prover trained on human proofs alone and almost as many as with the combination of both approaches.
Hence, we solve one of the road blocks on the way to open-ended learning of mathematical reasoning in large theories.

\section{Related work}
\label{sec:related}

\paragraph{Reinforcement learning.}
Reinforcement learning without imitation learning has been successful for computer games (cf.~\citet{mnih2013playing}) and it was demonstrated later in~\citet{silver2017masteringzero} that imitation learning is not necessary for complex games like Chess and Go.
For more complex games with much larger action spaces,
learning methods still rely on human imitation due to the exploration problem~(cf.~\citet{oriol2019alphastar}).
The question of exploration is well studied in reinforcement learning~\citep{houthooft2016vime,burda2018large}, but existing approaches such as $\epsilon$-greedy do not work for premise selection because of very large (practically infinite) action space.

\paragraph{Learning premise selection.}
Premise selection has been an active research topic in the domain of automated theorem proving~\citep{alama2014premise,kaliszyk2015learning,blanchette2016learning,wang2017premise}. Neural networks were first applied to premise selection for automated theorem proving in~\citet{alemi2016deepmath}. Other works have focused on tactic selection~\citep{huang2018gamepad} and tactic generation~\citep{yang2019learning}.
All of these works rely on human proof data.
\cite{kaliszyk2018reinforcement}, \citet{zomboricurriculum} and \cite{zombori2020prolog} are based on reinforcement learning, however they do not address the problem of premise selection, which is the hard part of exploration due to the unbounded repository of premises.

We use the HOList environment~\citep{bansal2019holist} in our experiments. 
Similar to HOList for HOL Light~\citep{Harrison96}, GamePad~\citep{huang2018gamepad}, CoqGym~\citep{yang2019learning}, TacticToe~\citep{gauthier2017tactictoe}, and Proverbot9001~\citep{proverbot2019} are machine learning environments for proof assistants Coq~\citep{coq} and HOL4~\citep{slind2008brief}.
We chose to use HOList because of the breadth of topics of mathematics in the dataset.
Additionally, HOList is already integrated into a reinforcement learning setup, which our approach relies on.

\section{Background}
\label{sec:background}
\paragraph{Theorem proving.}
In this work we build on proof assistants, which have been built to enable humans to write and then automatically check proofs.
In contrast to mathematical textbooks and papers, which are written mostly in natural language, we call mathematics formalized in proof assistants to be \emph{formal mathematics}.
In this work we focus on the proof assistant HOL Light~\citep{Harrison96}, in which a wide range of mathematical theories have been formalized, and which has been famously used for the formalization of the Kepler conjecture~\citep{hales2017formal}.
Other proof assistants include Coq~\citep{coq}, HOL4~\citep{slind2008brief}, Isabelle~\citep{wenzel08isabelle}, and Lean~\citep{de2015lean}.
HOL Light, as in many other proof assistants, relies mostly on ``backward'' proof steps.
In contrast to ``forward'' proof steps, in which we only manipulate already proven statements, backward proofs start with a proof goal (the statement of the theorem to be proven) and apply proof tactics until all goals are proven.
\begin{figure}[b]
\centering
\begin{tikzpicture}[thick, >=latex,
pre/.style={<-,shorten >= 1pt, shorten <=1pt, thick},
post/.style={->,shorten >= 1pt, shorten <=1pt,  thick},
und/.style={very thick, draw=gray},
node1/.style={rectangle, minimum size=4mm, draw=black!100, line width=1pt, inner sep=2mm},
node2/.style={rectangle, minimum size=5mm, draw=white!100, fill=white!100, very thick, inner sep=1mm},
dummy/.style={rectangle, minimum size=0mm, draw=black!100,
very thick, inner sep=0},
virt/.style={circle,draw=black!50,fill=black!20, opacity=0}]

\newcommand{\xd}{0}
\newcommand{\yd}{0}
\newcommand{\xstep}{2}
\newcommand{\ystep}{1}

\node[node1] (s1) at (\xd, \yd+0.2*\ystep)
{$\forall x \in \Nats : x + 0 = x$};

\node[dummy] (d1) at (\xd, \yd-1*\ystep) {};

\node[dummy] (d2) at (\xd+2.25*\xstep, \yd-1*\ystep) {};

\node[node2] (t1) at (\xd+1.3*\xstep, \yd-0.75*\ystep)
{\textcolor{\myblue}
{\texttt{MATCH\_MP\_TAC}~~ \texttt{NAT\_INDUCTION}}};

\node[node1] (s2) at (\xd+3.64*\xstep, \yd-1*\ystep)
{$0 + 0 = 0$};

\node[node1] (s3) at (\xd+2.25*\xstep, \yd-2*\ystep)
{$\forall x \in \Nats : ((x + 0 = x) \,\Rightarrow\, (x + 1 + 0 = x + 1))$};

\node[node2] (t2) at (\xd+4.7*\xstep, \yd-0.75*\ystep)
{\textcolor{\myblue}{\texttt{ARITH\_TAC}}};

\node[node2] (s2p) at (\xd+5.7*\xstep, \yd-1*\ystep)
{\textcolor{\myred}{PROVEN}};

\node[node2] (t3) at (\xd+4.7*\xstep, \yd-1.75*\ystep)
{\textcolor{\myblue}{\texttt{ARITH\_TAC}}};

\node[node2] (s3p) at (\xd+5.7*\xstep, \yd-2*\ystep)
{\textcolor{\myred}{PROVEN}};

\draw[->, very thick] (s1) to (d1) to (s2);
\draw[->, very thick] (d2) to (s3);
\draw[->, very thick] (s2) to (s2p);
\draw[->, very thick] (s3) to (s3p);

\end{tikzpicture}
\caption{Formally proving $\forall x \in \Nats : x + 0 = x$.}
\label{fig:tp_example}
\end{figure}
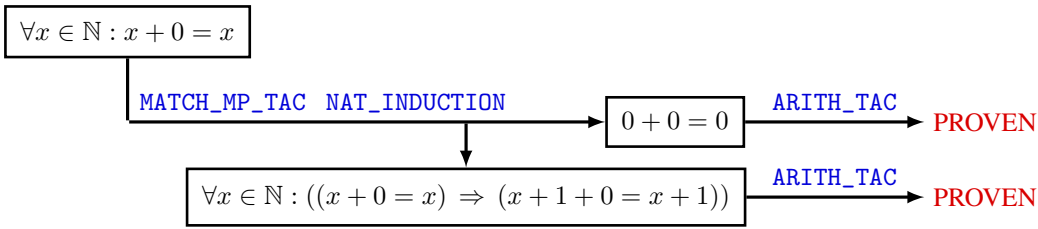

In Figure~\ref{fig:tp_example}, we give an example of a backward proof.
The goal here is to prove $x + 0 = x$, for all $x\in\mathbb{N}$, and we apply the tactic \texttt{MATCH\_MP\_TAC} to the goal.
Like many tactics, this tactic takes a \emph{premise} (i.e. a previously proven theorem or lemma) as a parameter.
In this example, we use the induction theorem \texttt{NAT\_INDUCTION} as a premise.
This tactic application splits the first goal into two subgoals, corresponding to the base case and the induction step.
The semantics of an application of a proof tactic is that, if all subgoals are proven, then also the goal to which the tactic has been applied is proven.
In our case, we can prove both of the subgoals by simple arithmetic reasoning, provided by the tactic \texttt{ARITH\_TAC}.
This tactic here does not require additional premises and returns an empty list of subgoals (for both of the subgoals we apply it to), meaning that they are proven, and hence the original goal is proven.

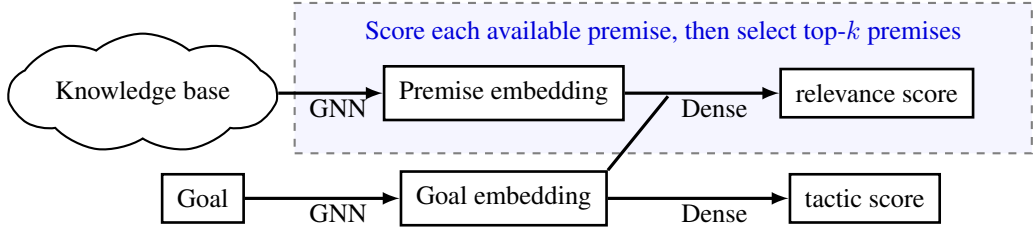
\begin{figure}
\centering
\begin{tikzpicture}[thick, >=latex,
pre/.style={<-,shorten >= 1pt, shorten <=1pt, thick},
post/.style={->,shorten >= 1pt, shorten <=1pt,  thick},
und/.style={very thick, draw=gray},
node1/.style={rectangle, minimum size=4mm, draw=black!100, fill=white!100, line width=1pt, inner sep=2mm},
node2/.style={rectangle, minimum size=5mm, opacity=.0,text opacity=1, very thick, inner sep=1mm},
dummy/.style={rectangle, minimum size=0mm, draw=black!100,
very thick, inner sep=0},
virt/.style={circle,draw=black!50,fill=black!20, opacity=0},
outer/.style={draw=gray,dashed,fill=blue!4,thick,inner sep=5pt}]

\newcommand{\xd}{0}
\newcommand{\yd}{0}
\newcommand{\xstep}{2}
\newcommand{\ystep}{1}
\newcommand{\yprem}{0.46}
\newcommand{\yprembeg}{2.3*\ystep}

\node[outer, minimum width=98mm, minimum height=20mm] (background)
at (\xd+3.06*\xstep, \yd+1.6*\ystep) {};
\node[node2] at (\xd+3.06*\xstep, \yd+2.2*\ystep)
{\textcolor{\myblue}{
Score each available premise, then select top-$k$ premises}};

\node[node1] (g) at (\xd, \yd)
{Goal};

\node[node2] (gnn1) at (\xd+0.9*\xstep, \yd-0.2*\ystep)
{GNN};
\node[node1] (gemb) at (\xd+2*\xstep, \yd)
{Goal embedding};

\node[node2] (gnn2) at (\xd+0.9*\xstep, \yd+1.2*\ystep)
{GNN};
\node[node1] (pemb) at (\xd+2*\xstep, \yd+1.4*\ystep)
{Premise embedding};

\draw[->, very thick] (g) to (gemb);

\node [cloud, draw, cloud puffs=9, cloud puff arc=80, aspect=3, inner ysep=1mm, minimum width=2mm]
(kb) at (\xd-0.4*\xstep, \yd+1.4*\ystep)
{Knowledge base};

\draw[->, very thick] (kb) to (pemb);

\node[node2] (de1) at (\xd+3.4*\xstep, \yd-0.2*\ystep)
{Dense};
\node[node2] (de2) at (\xd+3.4*\xstep, \yd+1.2*\ystep)
{Dense};
\node[node1] (tactic) at (\xd+4.4*\xstep, \yd)
{tactic score};

\draw[->, very thick] (gemb) to (tactic);

\node[node1] (pscore) at (\xd+4.48*\xstep, \yd+1.4*\ystep)
{relevance score};

\draw[->, very thick] (pemb) to (pscore);
\node[dummy] (d1) at (\xd+3.1*\xstep, \yd+1.38*\ystep){};
\draw[very thick] (gemb.north east) to (d1);

\end{tikzpicture}
\caption{Architecture for tactic and premise selection by~\cite{paliwal2020graph}. Note that this work is largely agnostic to the model architectures.}
\label{fig:nn}
\vspace{-2mm}
\end{figure}

\paragraph{Learning proof guidance for interactive theorem proving.}

It is a long-standing goal in artificial intelligence to automate the theorem proving process described above, in particular to relieve the human experts from selecting the tactics and premises in each proof step.
Historically, most works focused on designing advanced search algorithms, leading to entire fields such as SAT and SMT solving and first-order theorem proving.
Recently, learning proof search strategies from data has become an area of active research~\citep{alemi2016deepmath, gauthier2017tactictoe,loos2017deep,kaliszyk2018reinforcement,huang2018gamepad,bansal2019holist,lederman2020learningQBF}.

In this work, we follow the approach by~\cite{bansal2019holist}, which has shown the unique ability to find relevant premises, which has been a big challenge for the classical techniques. %
Figure~\ref{fig:nn} illustrates the tactic and premise selection architecture introduced by~\cite{bansal2019holist} and improved by~\cite{paliwal2020graph}.
For each proof step, this architecture scores the tactics and it also produces a relevance score for each potential premise.
While there are only few proof tactics, there are on average around 10,000 potential premises for each proof step.
The number of potential premises will likely further grow in the future if this approach is applied to even larger mathematical theories.
Then, for each tactic, the top-$k$ premises are given as arguments (unless the tactic does not require premise arguments).
This results in a list of candidate tactic applications, which can be used as the actions in any search approach.
We adopted the same search strategy as \cite{bansal2019holist} and \cite{paliwal2020graph}, which is a simple breadth-first search with a parameter of how many of the candidate tactic applications should be expanded per proof goal.

The tactic and premise selection architecture is trained on successful proofs.
For imitation learning, tuples of goal, tactic, and used premises are extracted from human proofs formalized in HOL Light.
The focus of this work, however, is to not learn from human proofs, and instead learn in a reinforcement learning setup from the proofs that earlier versions of the policy have found.

\section{Learning without Imitation}
\label{sec:learning}

In this section, we explain the setup for learning to prove in the absence of human proofs, and the considerations that informed our final design.

Much of mathematics that has been formalized by humans is in pursuit of formalizing certain theorems such as the four color theorem~\citep{gonthier2008formal} and the Kepler conjecture~\citep{hales2017formal}.
Since formalization is a challenging and tedious process requiring experts, a very small fraction of mathematics is formalized after decades of human effort.
This work paves the way for a critical piece of a system
wherein its knowledge base of formally proven theorems grows continuously.
We separate two key aspects of such a system.
First, proposing potentially true statements, also referred to in the literature as conjecturing.
Second, given a new statement, proving it in absence of existing proofs to learn from.
We would like to tackle the latter question in this work directly.

The key information the human proofs provide is the overall
direction of the proof via selection of the relevant premises at each proof step.
In case human proofs are available, one can first train
a machine learning model to imitate them
(as described in Section~\ref{sec:background}), as has been done in several previous works (Section~\ref{sec:related}).
\begin{figure}
\begin{subfigure}[t]{0.23\textwidth}
    \includegraphics[width=\textwidth]{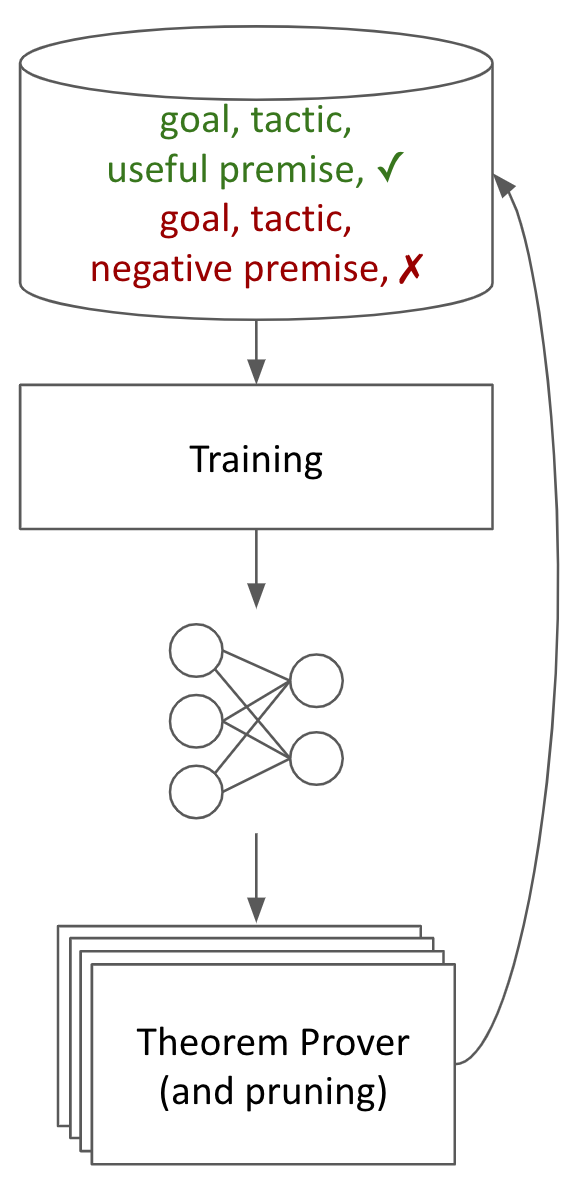}
    \caption{RL loop}
    \label{fig:rlloop}
\end{subfigure}
\hfill
\begin{subfigure}[t]{0.73\textwidth}
    \centering
    \includegraphics[width=\textwidth]{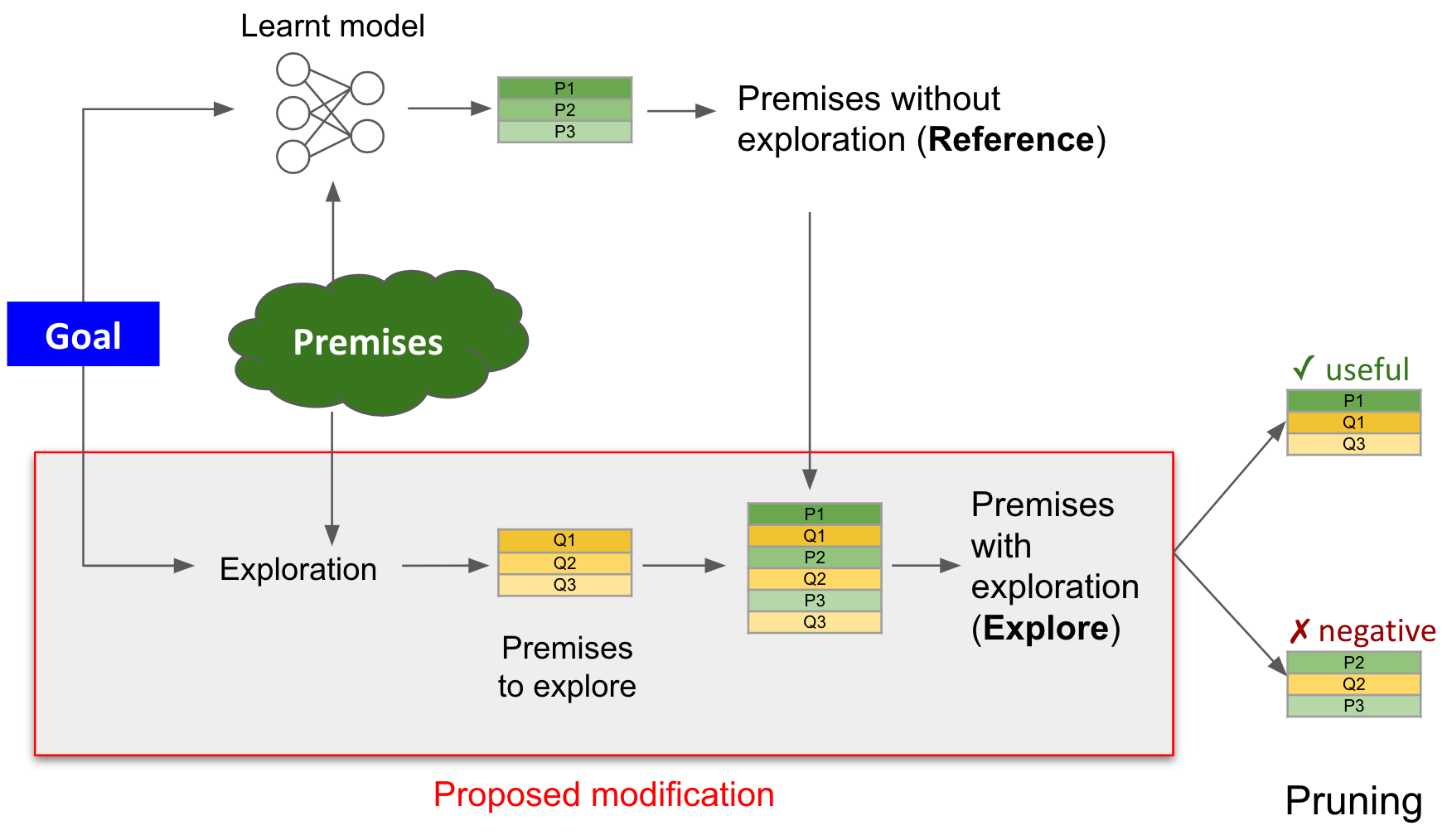}
    \caption{Change to the theorem prover to add exploration of premises.}
    \label{fig:explorepremises}
\end{subfigure}
\caption{The figure on the left gives a high-level overview of the components in the reinforcement learning (RL) loop. The figure on the right shows how the model being trained is used for premise selection. We propose a modification to the premise selection process to aid exploration in the RL loop.}
\label{fig:rloverview}
\vspace{-4mm}
\end{figure}
\paragraph{Reinforcement learning loop.}
In the absence of human proofs, we need a mechanism to incrementally improve a proof guidance model, which motivates the reinforcement learning setup we use.
Figure~\ref{fig:rlloop} shows the components of the reinforcement learning loop from \cite{bansal2019holist}, which we build upon.
A proof guidance model is trained with continuously expanding training data.
In order to generate the continuously expanding training data, several theorem provers run in lockstep with training the policy and premise selection network.
The provers try to prove the statements in the training set using the model for proof guidance as it is training.
If it manages to prove a statement, the proof is used to generate additional training data.
Intuitively, we would like to reward the choices of tactics and premises that were ``useful'' at each step in the proof.
A subtle but crucial aspect is that the proofs are pruned before generating the training data.
In this step, premises that are not necessary for the proof are removed. 
This interplay between over-approximation and pruning is a major contributing factor to the efficiency of our exploration method and is studied in Subsection~\ref{sec:overapproximation}.

Figure~\ref{fig:explorepremises} shows how the list of premises are picked, including our proposed change to aid exploration of the premises.
Given a goal, the currently learnt model is used to pick the top-$k_1$ highest scored premises from the knowledge base of premises available: $\{P_1, P_2, \ldots, P_{k_1}\}$.
Simultaneously, we propose generating another list of premises $\{Q_1, Q_2, \ldots Q_{k_2}\}$ to explore, picked according to a metric discussed shortly (Section \ref{sec:retrieval}).
The final set of premises is obtained by interleaving the two premise lists.
$k_1$ and $k_2$ are hyperparameters which can be varied to control exploitation of the learnt model (higher $k_1$) vs exploration (higher $k_2$). The \textsc{Reference} setup is one without our modification
(i.e., $k_2=0$).
On the other hand, the \textsc{Explore} setup includes additional premises as proposed here.

There are three aspects to this that we wish to highlight, and that have informed the design: the strategy to generate the list of premises to explore, the effect of using irrelevant premises in an action (over-approximation of premise lists), and pruning. We discuss each of these in detail, designing experiments to inform different choices.

\subsection{Information retrieval for premise selection}
\label{sec:retrieval}
One of the key failure modes of our reinforcement learning setup trained without human proofs is not being able to prove new theorems.
With no new training data, the learning process would stall.
Thus, it is crucial that we continuously expand the horizon of theorems that we are able to prove.
Since we generate new training data only when we manage to prove a theorem, we wish to pick premises most likely relevant to prove the current goal.

In information retrieval literature, notions such as
tf-idf~\citep{manning2008introduction} have been used to retrieve relevant documents corresponding to a query.
We view premise selection as a retrieval problem, thinking of the current goal we are trying to prove as the query, and the knowledge base of previously proven theorems (premises) as the documents from which we would like to retrieve.

Given a goal $G$, we use pre-engineered similarity
scoring $s(G, P)$ to rank a potential premise $P$ for its usefulness
in the next action (tactic application) with the target of proving the goal.
In our setup, we restrict our attention to functions of the form 
$s(G, P)=\langle r(G)/\|r(G)\|, r(P)/\|r(P)\|\rangle$,
where $r$
is some simple vector representation of the expression and $\langle \cdot, \cdot \rangle$ the dot product. This is also sometimes referred to as the cosine simarity.
We consider $r$ of the form $r(P)_i=\mbox{tf}(P, i)\mbox{idf}(i)$, where the $i$-component corresponds to the tokens occurring in the formulas, $\mbox{idf}$ is the ``inverse document frequency'' function which is precomputed by $\mbox{idf}(i) = \log(N/n_i),$ where $N$ is the total number of theorems and $n_i$ is the number of theorem containing the $i$-th token. For the term frequency function $\mbox{tf}(P, i)$, we have tested three possible choices: boolean weighting: $1$ if $f_i(P)>0$ and $0$ otherwise, logarithm weighting: $1+\log(f_i(P))$ and natural: $\mbox{tf}(P, i)=f_i(P)$, where $f_i(P)$ is the number of occurrences of the $i$-th token in expression $P$. 

Running a full reinforcement learning loop uses over 25 years of CPU resources (see supplement, Appendix~\ref{apx:loopresources}).
Since it is prohibitive to run a lot of experiments with the full system, it suggests that we should first evaluate the quality of similarity metrics in an offline manner.
That is, we measure how well those metrics perform on existing proof traces and we only apply the similarity scores that performed best in a separate evaluation.

\begin{table}
    \centering
    \caption{Retrieval performance on human proof logs}
    \begin{tabular}{|c|c|c|c|c|c|}
    \hline
         Term Freq. & av rel max rank & recall@16 & recall@32 & recall@64 & recall@128 \\ \hline
         boolean & 0.24 & 0.15 & 0.19 & 0.25 & 0.31 \\ \hline
         logarithm & 0.35 & 0.1 & 0.13 & 0.17 & 0.21 \\ \hline
         natural & 0.46 & 0.06 & 0.08 & 0.09 & 0.11 \\ \hline
    \end{tabular}
    \label{table:retrieval}
\end{table}
In order to measure the quality of a similarity score, we have adopted
the metrics from~\citet{alemi2016deepmath}. We have measured
the average relative maximum rank and the top-$k$ recall numbers for a small
set of relevant $k$ values (for $k=8,16,32$ and $64$) on a random selection of proofs comprising of $20\%$ of the training set of the ``complex'' corpus of HOList.
The relative maximum rank of a true positive document (premise) is the absolute maximum rank of the true documents divided by the size of all possible premises (from which we make the selection), the average is then taken over all retrieval tasks.
The results are summarized in Table~\ref{table:retrieval}. Note that low average maximum relative rank and high recall values indicate better premise selection performance of the similarity measure.

To summarize, we pick the \textbf{boolean} term-frequency weighting scheme. In addition, we speculate it could be helpful to add more variation of premises picked for exploration for a given goal, as over the course of a loop same goals are attempted multiple times. To add this variation, we add a dropout probability hyperparameter $p$ to components of the representation: $r(G)_i$ is zeroed with probability $p$ when computing the representation for $G$. $p$ is set to $0.1$ unless specified otherwise.

\subsection{Over-approximation of premises}
\label{sec:overapproximation}
Our exploration is based on the assumption that adding a few irrelevant premises does not influence the outcome of tactic applications significantly. This allows us to accelerate the search by trying to over-approximate the set of premises used in tactic applications.
We study the behavior of the most frequently occurring tactics that take premises as an argument: {\tt MESON} and {\tt REWRITE}.

{\tt MESON} is based on a first-order logic solver. We study how many extra premises we can add before {\tt MESON} starts to fail to prove a goal.
To do so, we first sample at random proof steps from the human proofs (which are always successful).
For each proof step we add as tactic arguments random irrelevant premises from the
knowledge base of
premises available at that proof step.
We report the ratio of successful proof attempts after adding a set of tactic parameters with varying cardinality.
For {\tt REWRITE} operations, we study the number of extra premises we can add and expect not to change the outcome of the rewrite operation. 
For both experiments, we sampled random proofs and one random
proof step with the desired type of tactic ({\tt MESON} or {\tt REWRITE}), until we sampled 250
steps successfully. Then for each $l\in\{1,2,4,8,16,32\},$ we sampled five different random parameter lists with length $l$. In our experiment, we append those parameters to our parameter list and execute the same tactic with the extended parameter list. Table~\ref{table:tacticstats} shows the ratio of application with the outcome being identical with that of the tactic application without the extra parameters. We can see that even adding 32 extra random parameters does not change the outcome of the rewrite tactics over 70\% of the time.
However, {\tt MESON} tends to time out with more than 32 extra premises.
\begin{table}
\begin{minipage}{\linewidth} \centering
\caption{Tactic success rates with extra random parameters, 1 second timeout.}
\label{table:tacticstats}
\begin{tabular}{|c|c|c|c|c|c|c|}
\hline
Number of extra premises & 1 & 2 & 4 & 8 & 16 & 32 \\
\hline
{\tt MESON} success rate  & 0.995 & 0.986 & 0.97 & 0.873 & 0.53 & 0.06 \\
\hline
{\tt REWRITE} unchanged rate & 
0.99 & 0.979 & 0.954 & 0.93 & 0.858 & 0.731 \\
\hline
\end{tabular}
\end{minipage}
\end{table}
\subsection{Pruning}
\label{sec:pruning}
As discussed in Section \ref{sec:overapproximation}, if a tactic application succeeds, not all premises provided might have been used. In fact, we are using this fact to accelerate our exploration.
However, we do not wish to learn from these irrelevant premises. Thus, 
for generating training data, for each proof step in our proof we greedily try to remove the premises: given a proof step with premises
$\{P_i\}_{i=1}^{n}$, we rerun the proof step without $P_n$.
If the result of the proof step remains unchanged, we drop $P_n$. Then, we continue with trying to drop $P_{n-1}$, and so on.
We use the dropped premises as hard negatives, such that premises which are ranked highly by the model but are not useful are demoted.
Demoting pruned premises allows other premises that had a high score but did not make it into the top-$k$ to get a chance in the future.
Pruning also ensures that any extra premises added for exploration that are in fact unnecessary are not learnt upon.

\section{Evaluation}
\label{sec:results}
\paragraph{Environment and benchmark.}
We evaluate our approach in the HOList environment~\citep{bansal2019holist} based on the HOL Light~\citep{Harrison96} theorem prover.
We conduct our experiments on the ``complex'' corpus of the HOList benchmark derived from theorems in HOL Light's mathematical library from various areas of mathematics such as topology, multivariate calculus, real and complex analysis, geometric algebra, and measure theory.
It includes well-known theorems such as Abel's theorem for power series, the fundamental theorem of calculus, and that the roots of the characteristic polynomial of a complex matrix are its eigenvalues. Table \ref{table:holistexamples} gives two more concrete examples of theorems in the benchmark.

The task is to attempt to prove a theorem in the benchmark, with theorems and definitions
appearing before it in the benchmark available as premises.
The evaluation criterion is the fraction of theorems proven with this constraint.
The benchmark comes with theorems divided into training, validation, and test sets.
Table \ref{table:holiststats} gives the statistics of theorems in the ``complex'' corpus of the benchmark, which we attempt to prove. Definitions (totaling 637) and ``core'' corpus of theorems (totaling 2320), containing basic mathematics which the ``complex'' corpus builds upon, are available as premises, but are not attempted to be proven.

\begin{table}[b]
\begin{minipage}{.6\linewidth} \centering
\caption{Examples of theorems in the benchmark (compressed for brevity)}
\label{table:holistexamples}
\begin{tabular}{|c|}
\hline
  Alternative characterization of orthogonal matrices.\\
  $\forall A. orth(A) \iff (\forall i. \left|A_i\right|=1 \wedge \forall i != j. A_i \perp A_j)$ \\
\hline
  Property about absolute neighborhood retract (ANR).\\
  $\forall S \subseteq R^n.
  \operatorname{ANR} (\operatorname{frontier}(S)) \implies
  \operatorname{ANR} (\operatorname{closure}(S))$ \\
\hline
\end{tabular}
\end{minipage}
\hspace{0.05\linewidth}
\begin{minipage}{.3\linewidth} \centering
\caption{Benchmark statistics}
\label{table:holiststats}
\begin{tabular}{|c|c|}
\hline
Split      & \# of Theorems \\
\hline
Training   & 10214 \\
Validation & 3225 \\
Testing    & 3184 \\
\hline
Total      & 16623 \\
\hline
\end{tabular}
\end{minipage}
\end{table}

\paragraph{Training and evaluation.}
During training, we generate data for training by trying to prove statements in the
training set of 10,214 theorems.
We train for 8 million steps.
Details of our hardware
setup and hyperparameters are provided in the supplementary materials.
For evaluation of all our experiments trained in the reinforcement learning setup, we focus on the
number of statements proven in the held-out validation set of 3,225 theorems.
We run a continuous evaluation
on samples of the validation set as well as a final evaluation on the full validation set.
The precise meaning of the metrics in our plots and tables are given below.
\begin{itemize}[itemsep=0mm,leftmargin=4mm]
    \item \textbf{Continuous} validation performance (represented by dots in the plots) runs every 80,000 training steps on
    a \emph{random sample} of validation theorems, and reports the fraction of proven theorems from that sample.
    Since not all validation theorems are attempted and the sample changes each evaluation,
    the metric is slightly noisy, but it allows us to monitor overfitting during training.
    \item \textbf{Final} validation performance (reported in the tables and plots) is the fraction of all validation
    theorems proven by the final checkpoint at 8 million steps.
    This metric also allows for comparison with models trained purely by imitation learning.
    \item \textbf{Cumulative} validation
    performance -- reported in the tables and plots -- is the fraction of all validation
    theorems proven by any continuous-validation
    run up until that point in the loop. The table reports the cumulative
    performance of the whole loop (i.e., after 8 million steps).
\end{itemize}

\paragraph{Main experiment.}

\begin{figure}
\centering
\begin{minipage}{0.49\textwidth}
\footnotesize
\begin{tabular}{l|c|c}
                            & Final      & Cumulative\\
                            & validation & validation\\ 
\hline
\emph{Pure human imitation} & & \\
\quad \citet{bansal2019holist}    & 32.65\%    &  - \\
\quad \citet{paliwal2020graph}    & \textbf{49.95}\%    &  - \\
\hline
\emph{Human RL}    & & \\
\quad \citet{bansal2019holist}    & 38.9\%     & not reported \\
\quad Human reference             & 59.5\%    & 68.2\%\\
\quad Human explore               & \textbf{59.9}\%     & 69.1\% \\
\hline
\emph{Zero RL}      & & \\
\quad Zero reference              & 7.0\%      & 7.3\% \\
\quad Zero explore                & \textbf{56.3}\%     & 64.2\% \\
\end{tabular}
\end{minipage}
\begin{minipage}{0.49\textwidth}\centering
    \includegraphics[height=2in]{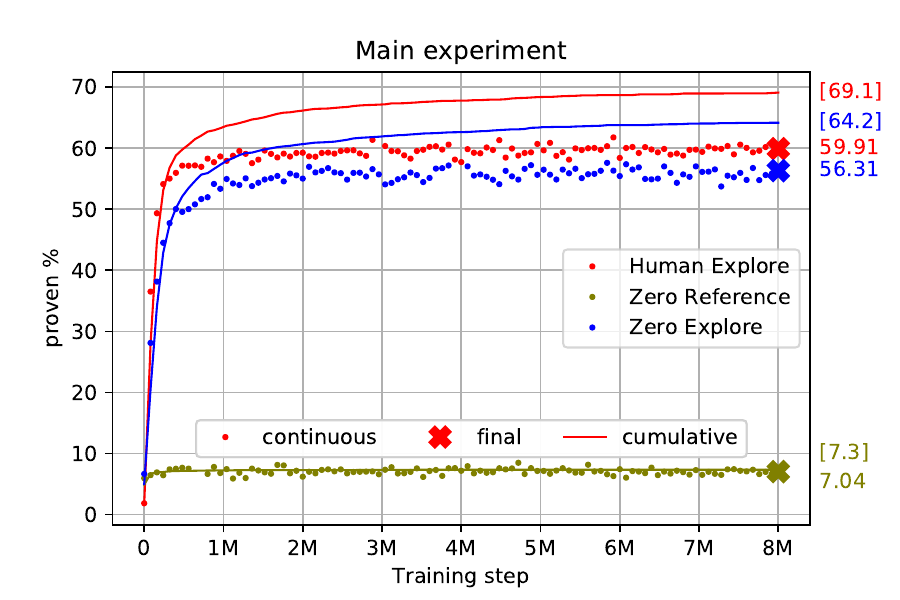}
\end{minipage}
\vspace{-2mm}
\caption{Results of our main experiment. We report the percentage of validation theorems proven on the HOList benchmark. The numbers in bold are the state-of-the-art in their respective categories (including results from this work). The main takeaway is that the best RL loop trained without human proof data outperforms the model trained purely on human data, and approaches the performance of the best RL loop trained with human data.}
\label{fig:main}
\vspace{-3mm}
\end{figure}In the main experiment we are interested in understanding the ability of the \textsc{Explore} approach proposed in Section \ref{sec:learning}, in particular, its ability to learn to prove without human proof data available. This experiment is referred to as \textsc{Zero Explore}.
As we see in the plot on the right in Figure~\ref{fig:main}, the loop continuously expands the horizon of the theorems it is able to prove.

To put the performance in context, we categorize the results on this benchmark into three categories. First, \emph{pure human imitation}, wherein the model is trained only on human proof data and no theorem prover is used during training.
Second, \emph{human RL}, wherein the the model is trained
on human proof data as well data generated by running
the prover in a reinforcement learning (RL) loop.
Finally, \emph{zero RL}, wherein no human data is available, and all data is generated by running a prover in an RL loop.
The results are summarized in the table in Figure~\ref{fig:main}.

\citet{paliwal2020graph} based on graph neural networks (GNN) is the state-of-the-art on pure human imitation on this benchmark,
and we use the network architecture in our experiments as well.
Compared to pure human imitation, the \textsc{Zero Explore} RL loop using no human proofs is able to prove more theorems on a single checkpoint: 49.95\% (pure imitation) vs 56.3\% (zero RL).

Next, we compare to the best human RL loop (\textsc{Human Explore}), one with everything identical as in \textsc{Zero Explore}, except we let the model learn from additional human proof data.
We see that the \textsc{Zero Explore} loop comes very close to the performance of the corresponding human RL loop: 59.9\% vs 56.3\% on a single checkpoint, and 69.1\% vs 64.1\% cumulatively.
\textsc{Zero Explore} is able to reach over 90\% of the human RL loop's performance.
This is an important indicator as \textsc{Human} loops indicate a ceiling unrelated to availability of proof data: such as the proof search algorithm, or model architecture.%

Finally, we compare against the RL setup from~\citet{bansal2019holist}, and run it without human proof data (\textsc{Zero Reference}). We run into the failure mode of not being able to prove new statements and thus stalling, discussed in Section \ref{sec:retrieval}.
\section{Ablation studies}
\begin{figure}
    \centering
    \begin{subfigure}[b]{0.5\textwidth}
        \includegraphics[height=1.8in]{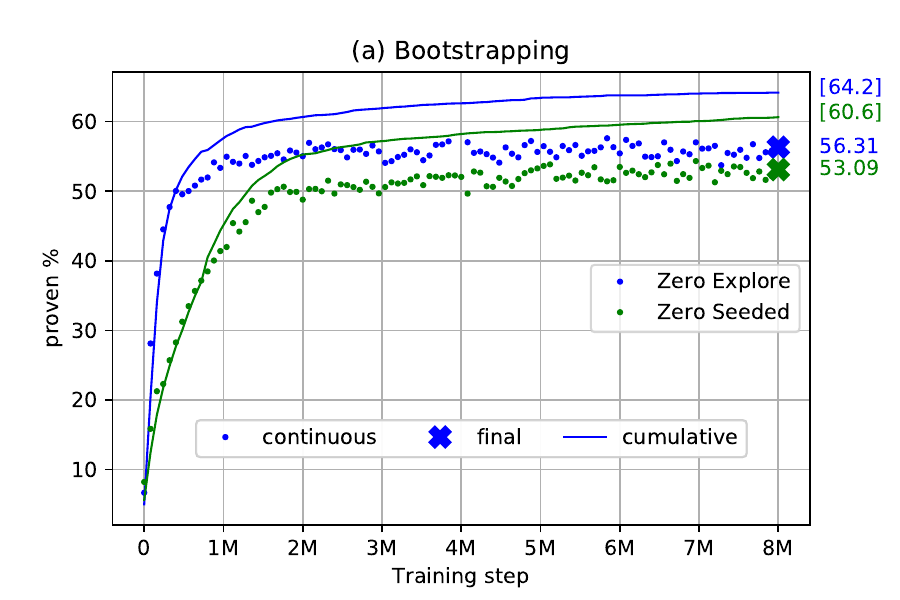}
    \end{subfigure}
    \begin{subfigure}[b]{0.49\textwidth}
        \includegraphics[height=1.8in]{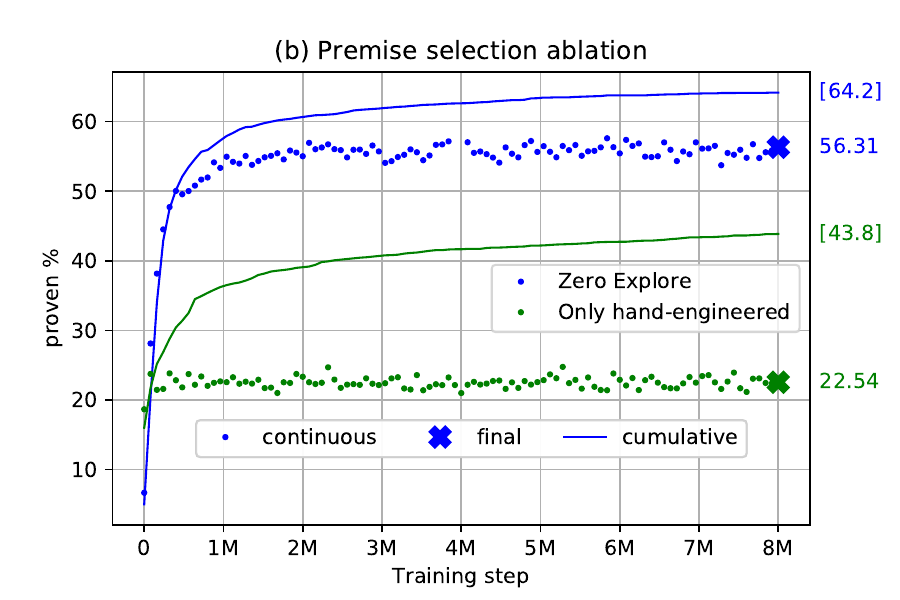}
    \end{subfigure}
    \begin{subfigure}[b]{0.5\textwidth}
        \includegraphics[height=1.8in]{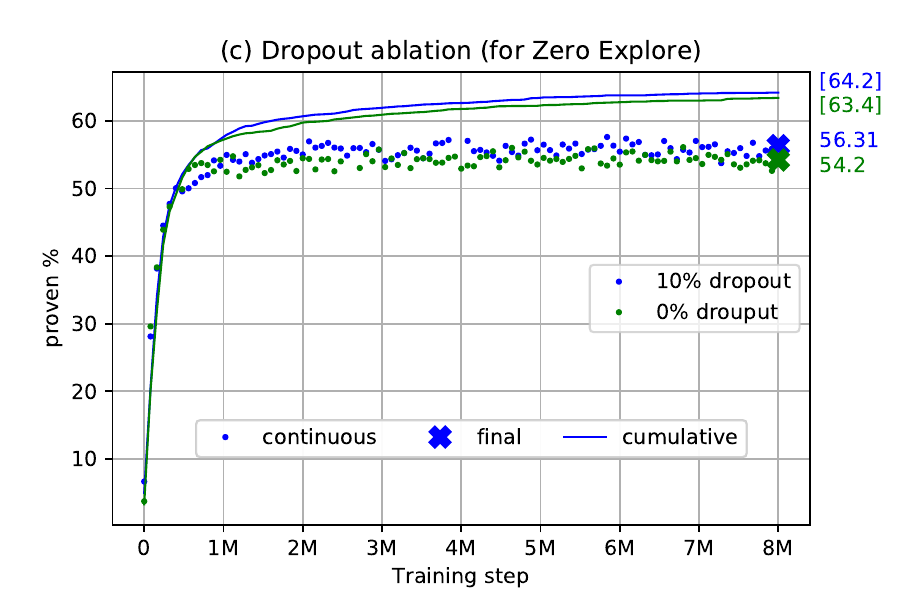}
    \end{subfigure}
    \begin{subfigure}[b]{0.49\textwidth}
        \includegraphics[height=1.8in]{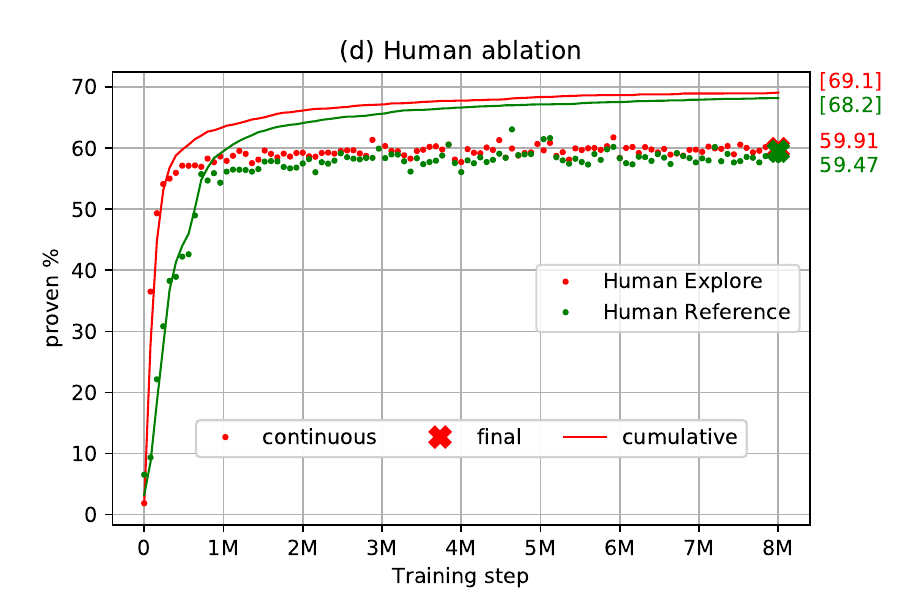}
    \end{subfigure}
\caption{Ablation experiments.
}
\label{fig:ablation}
\vspace{-4mm}
\end{figure}

\paragraph{Bootstrapping.}
In Section~\ref{sec:results}, we observe that the \textsc{Zero Reference} loop stalls very quickly. Here, we try to understand to what extent is the failure a bootstrapping issue.
We attempt to prove all statements in the training set using the best hand-engineered metric in Section~\ref{sec:retrieval} for premise selection and random tactic selection. This proves around 20\% percent of the statements.
We start a zero RL loop as in \textsc{Zero Reference}, but providing these additional proofs to train upon, calling it \textsc{Zero Seeded}.
We see in Figure \ref{fig:ablation}a that it does not stall like the reference loop.
At the same time, it does not reach the same level of performance as the \textsc{Zero Explore}, which explores throughout.

\vspace{-2mm}
\paragraph{Premise selection ablation.}
Through this ablation, we wish to understand the capability of the hand-engineered metrics in Section~\ref{sec:retrieval} to prove theorems.
To keep the focus on premise selection, we still learn the tactic selection, and use similar amounts of resources trying to prove statements as in one RL loop.
Technically, we do this by setting $k_1$ (number of premises chosen by the learnt network) to $0$, and rest of the setup as in the \textsc{Zero Explore} loop.
In Figure~\ref{fig:ablation}, under premise-selection ablation we see the results: it manages to prove 43\% of the statements cumulatively, compared with 64\% when learning with a combination of exploration and exploitation in the RL loop.

\vspace{-2mm}
\paragraph{Dropout.}
In Section~\ref{sec:retrieval} we suggest a 10\% token dropout probability when
deciding premises to explore to introduce more diversity of premises in the loop
overall at the cost of picking slightly less relevant premises at a specific point.
We evaluate this experimentally (dropout ablation in Figure~\ref{fig:ablation}),
but do not see a major difference: we can observe a very slight gain requiring further verification.

\vspace{-2mm}
\paragraph{Human loop ablation.}
We run a human loop where we do not add premises for exploration 
(human ablation, Figure~\ref{fig:ablation}).
We do not see a significant difference in the performance.
It is not surprising, as the human loop has proofs for all statements and is thus not reliant on premise exploration to find relevant premises, unlike the zero RL loops.

\section{Conclusion}
In this work, we demonstrate that it is possible to learn a premise selection model for theorem proving in the absence of human proofs.
We show that, on our benchmark, we exceed the performance of a network trained purely on human proofs, and approach the performance of the system that combines reinforcement learning with imitation learning.

\bibliography{main}

\begin{thebibliography}{30}
\providecommand{\natexlab}[1]{#1}
\providecommand{\url}[1]{\texttt{#1}}
\expandafter\ifx\csname urlstyle\endcsname\relax
  \providecommand{\doi}[1]{doi: #1}\else
  \providecommand{\doi}{doi: \begingroup \urlstyle{rm}\Url}\fi

\bibitem[Alemi et~al.(2016)Alemi, Chollet, E{\'{e}}n, Irving, Szegedy, and
  Urban]{alemi2016deepmath}
Alexander~A. Alemi, Fran{\c{c}}ois Chollet, Niklas E{\'{e}}n, Geoffrey Irving,
  Christian Szegedy, and Josef Urban.
\newblock Deepmath - deep sequence models for premise selection.
\newblock In Daniel~D. Lee, Masashi Sugiyama, Ulrike von Luxburg, Isabelle
  Guyon, and Roman Garnett, editors, \emph{Advances in Neural Information
  Processing Systems 29: Annual Conference on Neural Information Processing
  Systems 2016, December 5-10, 2016, Barcelona, Spain}, pages 2235--2243, 2016.
\newblock URL
  \url{http://papers.nips.cc/paper/6280-deepmath-deep-sequence-models-for-premise-selection}.

\bibitem[Gauthier et~al.(2017)Gauthier, Kaliszyk, and
  Urban]{gauthier2017tactictoe}
Thibault Gauthier, Cezary Kaliszyk, and Josef Urban.
\newblock {T}actic{T}oe: {L}earning to reason with {HOL4} tactics.
\newblock In Thomas Eiter and David Sands, editors, \emph{LPAR-21, 21st
  International Conference on Logic for Programming, Artificial Intelligence
  and Reasoning, Maun, Botswana, May 7-12, 2017}, volume~46 of \emph{EPiC
  Series in Computing}, pages 125--143. EasyChair, 2017.
\newblock URL \url{https://easychair.org/publications/volume/LPAR-21}.

\bibitem[Loos et~al.(2017)Loos, Irving, Szegedy, and Kaliszyk]{loos2017deep}
Sarah Loos, Geoffrey Irving, Christian Szegedy, and Cezary Kaliszyk.
\newblock Deep network guided proof search.
\newblock In Thomas Eiter and David Sands, editors, \emph{LPAR-21, 21st
  International Conference on Logic for Programming, Artificial Intelligence
  and Reasoning, Maun, Botswana, May 7-12, 2017}, volume~46 of \emph{EPiC
  Series in Computing}, pages 85--105. EasyChair, 2017.
\newblock URL \url{https://easychair.org/publications/paper/ND13}.

\bibitem[Huang et~al.(2019)Huang, Dhariwal, Song, and
  Sutskever]{huang2018gamepad}
Daniel Huang, Prafulla Dhariwal, Dawn Song, and Ilya Sutskever.
\newblock {G}ame{P}ad: {A} learning environment for theorem proving.
\newblock In \emph{7th International Conference on Learning Representations,
  {ICLR} 2019, New Orleans, LA, USA, May 6-9, 2019}. OpenReview.net, 2019.
\newblock URL \url{https://openreview.net/forum?id=r1xwKoR9Y7}.

\bibitem[Bansal et~al.(2019)Bansal, Loos, Rabe, Szegedy, and
  Wilcox]{bansal2019holist}
Kshitij Bansal, Sarah~M Loos, Markus~N Rabe, Christian Szegedy, and Stewart
  Wilcox.
\newblock {HOL}ist: {A}n environment for machine learning of higher-order
  theorem proving.
\newblock In Kamalika Chaudhuri and Ruslan Salakhutdinov, editors,
  \emph{Proceedings of the 36th International Conference on Machine Learning,
  {ICML} 2019, 9-15 June 2019, Long Beach, California, {USA}}, volume~97 of
  \emph{Proceedings of Machine Learning Research}, pages 454--463. {PMLR},
  2019.
\newblock URL \url{http://proceedings.mlr.press/v97/bansal19a/bansal19a.pdf}.

\bibitem[Paliwal et~al.(2020)Paliwal, Loos, Rabe, Bansal, and
  Szegedy]{paliwal2020graph}
Aditya Paliwal, Sarah Loos, Markus Rabe, Kshitij Bansal, and Christian Szegedy.
\newblock Graph representations for higher-order logic and theorem proving.
\newblock In \emph{The Thirty-Fourth {AAAI} Conference on Artificial
  Intelligence, {AAAI} 2020, New York, NY, USA, February 7-12, 2020}. {AAAI}
  Press, 2020.
\newblock ISBN 978-1-57735-823-7.
\newblock URL
  \url{https://www.aaai.org/Papers/AAAI/2020GB/AAAI-PaliwalA.9149.pdf}.

\bibitem[Lee et~al.(2020)Lee, Szegedy, Rabe, Loos, and
  Bansal]{lee2020mathematical}
Dennis Lee, Christian Szegedy, Markus~N. Rabe, Sarah~M. Loos, and Kshitij
  Bansal.
\newblock Mathematical reasoning in latent space.
\newblock In \emph{8th International Conference on Learning Representations,
  {ICLR} 2020, Addis Ababa, Ethiopia, April 26-30, 2020}. OpenReview.net, 2020.
\newblock URL \url{https://openreview.net/forum?id=Ske31kBtPr}.

\bibitem[Manning et~al.(2008)Manning, Raghavan, and
  Sch{\"{u}}tze]{manning2008introduction}
Christopher~D. Manning, Prabhakar Raghavan, and Hinrich Sch{\"{u}}tze.
\newblock \emph{Introduction to Information Retrieval}.
\newblock Cambridge University Press, 2008.
\newblock ISBN 978-0-521-86571-5.

\bibitem[Mnih et~al.(2013)Mnih, Kavukcuoglu, Silver, Graves, Antonoglou,
  Wierstra, and Riedmiller]{mnih2013playing}
Volodymyr Mnih, Koray Kavukcuoglu, David Silver, Alex Graves, Ioannis
  Antonoglou, Daan Wierstra, and Martin~A. Riedmiller.
\newblock Playing atari with deep reinforcement learning.
\newblock \emph{CoRR}, abs/1312.5602, 2013.
\newblock URL \url{http://arxiv.org/abs/1312.5602}.

\bibitem[Silver et~al.(2017)Silver, Schrittwieser, Simonyan, Antonoglou, Huang,
  Guez, Hubert, Baker, Lai, Bolton, et~al.]{silver2017masteringzero}
David Silver, Julian Schrittwieser, Karen Simonyan, Ioannis Antonoglou, Aja
  Huang, Arthur Guez, Thomas Hubert, Lucas Baker, Matthew Lai, Adrian Bolton,
  et~al.
\newblock Mastering the game of go without human knowledge.
\newblock \emph{Nature}, 550\penalty0 (7676):\penalty0 354, 2017.

\bibitem[Vinyals et~al.(2019)Vinyals, Babuschkin, Czarnecki, Mathieu, Dudzik,
  Chung, Choi, Powell, Ewalds, Georgiev, Oh, Horgan, Kroiss, Danihelka, Huang,
  Sifre, Cai, Agapiou, Jaderberg, Vezhnevets, Leblond, Pohlen, Dalibard,
  Budden, Sulsky, Molloy, Paine, Gulcehre, Wang, Pfaff, Wu, Ring, Yogatama,
  Wünsch, McKinney, Smith, Schaul, Lillicrap, Kavukcuoglu, Hassabis, Apps, and
  Silver]{oriol2019alphastar}
Oriol Vinyals, Igor Babuschkin, Wojciech~M Czarnecki, Michaël Mathieu, Andrew
  Dudzik, Junyoung Chung, David~H Choi, Richard Powell, Timo Ewalds, Petko
  Georgiev, Junhyuk Oh, Dan Horgan, Manuel Kroiss, Ivo Danihelka, Aja Huang,
  Laurent Sifre, Trevor Cai, John~P Agapiou, Max Jaderberg, Alexander~S
  Vezhnevets, Rémi Leblond, Tobias Pohlen, Valentin Dalibard, David Budden,
  Yury Sulsky, James Molloy, Tom~L Paine, Caglar Gulcehre, Ziyu Wang, Tobias
  Pfaff, Yuhuai Wu, Roman Ring, Dani Yogatama, Dario Wünsch, Katrina McKinney,
  Oliver Smith, Tom Schaul, Timothy Lillicrap, Koray Kavukcuoglu, Demis
  Hassabis, Chris Apps, and David Silver.
\newblock {Grandmaster level in StarCraft II using multi-agent reinforcement
  learning}.
\newblock \emph{Nature}, 575\penalty0 (7782):\penalty0 350--354, 2019.
\newblock ISSN 0028-0836.

\bibitem[Houthooft et~al.(2016)Houthooft, Chen, Duan, Schulman, Turck, and
  Abbeel]{houthooft2016vime}
Rein Houthooft, Xi~Chen, Yan Duan, John Schulman, Filip~De Turck, and Pieter
  Abbeel.
\newblock {VIME:} variational information maximizing exploration.
\newblock In Daniel~D. Lee, Masashi Sugiyama, Ulrike von Luxburg, Isabelle
  Guyon, and Roman Garnett, editors, \emph{Advances in Neural Information
  Processing Systems 29: Annual Conference on Neural Information Processing
  Systems 2016, December 5-10, 2016, Barcelona, Spain}, pages 1109--1117, 2016.
\newblock URL
  \url{http://papers.nips.cc/paper/6591-vime-variational-information-maximizing-exploration}.

\bibitem[Burda et~al.(2019)Burda, Edwards, Pathak, Storkey, Darrell, and
  Efros]{burda2018large}
Yuri Burda, Harrison Edwards, Deepak Pathak, Amos~J. Storkey, Trevor Darrell,
  and Alexei~A. Efros.
\newblock Large-scale study of curiosity-driven learning.
\newblock In \emph{7th International Conference on Learning Representations,
  {ICLR} 2019, New Orleans, LA, USA, May 6-9, 2019}. OpenReview.net, 2019.
\newblock URL \url{https://openreview.net/forum?id=rJNwDjAqYX}.

\bibitem[Alama et~al.(2014)Alama, Heskes, K{\"{u}}hlwein, Tsivtsivadze, and
  Urban]{alama2014premise}
Jesse Alama, Tom Heskes, Daniel K{\"{u}}hlwein, Evgeni Tsivtsivadze, and Josef
  Urban.
\newblock Premise selection for mathematics by corpus analysis and kernel
  methods.
\newblock \emph{J. Autom. Reasoning}, 52\penalty0 (2):\penalty0 191--213, 2014.
\newblock URL \url{https://doi.org/10.1007/s10817-013-9286-5}.

\bibitem[Kaliszyk and Urban(2015)]{kaliszyk2015learning}
Cezary Kaliszyk and Josef Urban.
\newblock Learning-assisted theorem proving with millions of lemmas.
\newblock \emph{J. Symb. Comput.}, 69:\penalty0 109--128, 2015.
\newblock URL \url{https://doi.org/10.1016/j.jsc.2014.09.032}.

\bibitem[Blanchette et~al.(2016)Blanchette, Greenaway, Kaliszyk,
  K{\"{u}}hlwein, and Urban]{blanchette2016learning}
Jasmin~Christian Blanchette, David Greenaway, Cezary Kaliszyk, Daniel
  K{\"{u}}hlwein, and Josef Urban.
\newblock A learning-based fact selector for {I}sabelle/{HOL}.
\newblock \emph{Journal of Automated Reasoning}, 57\penalty0 (3):\penalty0
  219--244, 2016.
\newblock URL \url{https://doi.org/10.1007/s10817-016-9362-8}.

\bibitem[Wang et~al.(2017)Wang, Tang, Wang, and Deng]{wang2017premise}
Mingzhe Wang, Yihe Tang, Jian Wang, and Jia Deng.
\newblock Premise selection for theorem proving by deep graph embedding.
\newblock In Isabelle Guyon, Ulrike von Luxburg, Samy Bengio, Hanna~M. Wallach,
  Rob Fergus, S.~V.~N. Vishwanathan, and Roman Garnett, editors, \emph{Advances
  in Neural Information Processing Systems 30: Annual Conference on Neural
  Information Processing Systems 2017, 4-9 December 2017, Long Beach, CA,
  {USA}}, pages 2786--2796, 2017.
\newblock URL
  \url{http://papers.nips.cc/paper/6871-premise-selection-for-theorem-proving-by-deep-graph-embedding}.

\bibitem[Yang and Deng(2019)]{yang2019learning}
Kaiyu Yang and Jia Deng.
\newblock Learning to prove theorems via interacting with proof assistants.
\newblock In Kamalika Chaudhuri and Ruslan Salakhutdinov, editors,
  \emph{Proceedings of the 36th International Conference on Machine Learning,
  {ICML} 2019, 9-15 June 2019, Long Beach, California, {USA}}, volume~97 of
  \emph{Proceedings of Machine Learning Research}, pages 6984--6994. {PMLR},
  2019.
\newblock URL \url{http://proceedings.mlr.press/v97/yang19a/yang19a.pdf}.

\bibitem[Kaliszyk et~al.(2018)Kaliszyk, Urban, Michalewski, and
  Ols{\'{a}}k]{kaliszyk2018reinforcement}
Cezary Kaliszyk, Josef Urban, Henryk Michalewski, and Miroslav Ols{\'{a}}k.
\newblock Reinforcement learning of theorem proving.
\newblock In Samy Bengio, Hanna~M. Wallach, Hugo Larochelle, Kristen Grauman,
  Nicol{\`{o}} Cesa{-}Bianchi, and Roman Garnett, editors, \emph{Advances in
  Neural Information Processing Systems 31: Annual Conference on Neural
  Information Processing Systems 2018, NeurIPS 2018, 3-8 December 2018,
  Montr{\'{e}}al, Canada}, pages 8836--8847, 2018.
\newblock URL
  \url{http://papers.nips.cc/paper/8098-reinforcement-learning-of-theorem-proving}.

\bibitem[Zombori et~al.(2019)Zombori, Csisz{\'{a}}rik, Michalewski, Kaliszyk,
  and Urban]{zomboricurriculum}
Zsolt Zombori, Adri{\'{a}}n Csisz{\'{a}}rik, Henryk Michalewski, Cezary
  Kaliszyk, and Josef Urban.
\newblock Towards finding longer proofs.
\newblock \emph{CoRR}, abs/1905.13100, 2019.
\newblock URL \url{http://arxiv.org/abs/1905.13100}.

\bibitem[Zombori et~al.(2020)Zombori, Urban, and Brown]{zombori2020prolog}
Zsolt Zombori, Josef Urban, and Chad~E. Brown.
\newblock Prolog technology reinforcement learning prover.
\newblock \emph{CoRR}, abs/2004.06997, 2020.
\newblock URL \url{https://arxiv.org/abs/2004.06997}.

\bibitem[Harrison(1996)]{Harrison96}
John Harrison.
\newblock {HOL L}ight: {A} tutorial introduction.
\newblock In Mandayam~K. Srivas and Albert~John Camilleri, editors,
  \emph{Formal Methods in Computer-Aided Design, First International
  Conference, {FMCAD} '96, Palo Alto, California, USA, November 6-8, 1996,
  Proceedings}, volume 1166 of \emph{Lecture Notes in Computer Science}, pages
  265--269. Springer, 1996.
\newblock URL \url{https://doi.org/10.1007/BFb0031795}.

\bibitem[Sanchez-Stern et~al.(2019)Sanchez-Stern, Alhessi, Saul, and
  Lerner]{proverbot2019}
Alex Sanchez-Stern, Yousef Alhessi, Lawrence Saul, and Sorin Lerner.
\newblock Generating correctness proofs with neural networks.
\newblock \emph{CoRR}, abs/1907.07794, 2019.
\newblock URL \url{http://arxiv.org/abs/1907.07794}.

\bibitem[Coq()]{coq}
Coq.
\newblock The {Coq Proof Assistant}, 1989.
\newblock URL \url{http://coq.inria.fr}.

\bibitem[Slind and Norrish(2008)]{slind2008brief}
Konrad Slind and Michael Norrish.
\newblock A brief overview of {HOL4}.
\newblock In Otmane~A{\"{\i}}t Mohamed, C{\'{e}}sar~A. Mu{\~{n}}oz, and
  Sofi{\`{e}}ne Tahar, editors, \emph{Theorem Proving in Higher Order Logics,
  21st International Conference, TPHOLs 2008, Montreal, Canada, August 18-21,
  2008. Proceedings}, volume 5170 of \emph{Lecture Notes in Computer Science},
  pages 28--32. Springer, 2008.
\newblock URL \url{https://doi.org/10.1007/978-3-540-71067-7\_6}.

\bibitem[Hales et~al.(2017)Hales, Adams, Bauer, Dang, Harrison, Le~Truong,
  Kaliszyk, Magron, McLaughlin, Nguyen, et~al.]{hales2017formal}
Thomas Hales, Mark Adams, Gertrud Bauer, Tat~Dat Dang, John Harrison, Hoang
  Le~Truong, Cezary Kaliszyk, Victor Magron, Sean McLaughlin, Tat~Thang Nguyen,
  et~al.
\newblock A formal proof of the kepler conjecture.
\newblock In \emph{Forum of Mathematics, Pi}, volume~5. Cambridge University
  Press, 2017.
\newblock URL
  \url{https://www.cambridge.org/core/services/aop-cambridge-core/content/view/78FBD5E1A3D1BCCB8E0D5B0C463C9FBC/S2050508617000014a.pdf/formal_proof_of_the_kepler_conjecture.pdf}.

\bibitem[Wenzel et~al.(2008)Wenzel, Paulson, and Nipkow]{wenzel08isabelle}
Makarius Wenzel, Lawrence~C. Paulson, and Tobias Nipkow.
\newblock The isabelle framework.
\newblock In Otmane~A{\"{\i}}t Mohamed, C{\'{e}}sar~A. Mu{\~{n}}oz, and
  Sofi{\`{e}}ne Tahar, editors, \emph{Theorem Proving in Higher Order Logics,
  21st International Conference, TPHOLs 2008, Montreal, Canada, August 18-21,
  2008. Proceedings}, volume 5170 of \emph{Lecture Notes in Computer Science},
  pages 33--38. Springer, 2008.
\newblock URL \url{https://doi.org/10.1007/978-3-540-71067-7\_7}.

\bibitem[de~Moura et~al.(2015)de~Moura, Kong, Avigad, Van~Doorn, and von
  Raumer]{de2015lean}
Leonardo de~Moura, Soonho Kong, Jeremy Avigad, Floris Van~Doorn, and Jakob von
  Raumer.
\newblock The {L}ean theorem prover (system description).
\newblock In Amy~P. Felty and Aart Middeldorp, editors, \emph{Automated
  Deduction - {CADE-25} - 25th International Conference on Automated Deduction,
  Berlin, Germany, August 1-7, 2015, Proceedings}, volume 9195 of \emph{Lecture
  Notes in Computer Science}, pages 378--388. Springer, 2015.
\newblock URL \url{https://doi.org/10.1007/978-3-319-21401-6}.

\bibitem[Lederman et~al.(2020)Lederman, Rabe, Seshia, and
  Lee]{lederman2020learningQBF}
Gil Lederman, Markus~N. Rabe, Sanjit Seshia, and Edward~A. Lee.
\newblock Learning heuristics for quantified boolean formulas through
  reinforcement learning.
\newblock In \emph{8th International Conference on Learning Representations,
  {ICLR} 2020, Addis Ababa, Ethiopia, April 26-30, 2020}. OpenReview.net, 2020.
\newblock URL \url{https://openreview.net/forum?id=BJluxREKDB}.

\bibitem[Gonthier(2008)]{gonthier2008formal}
Georges Gonthier.
\newblock Formal proof--the four-color theorem.
\newblock \emph{Notices of the AMS}, 55\penalty0 (11):\penalty0 1382--1393,
  2008.

\end{thebibliography}

\newpage

\appendix

\section{Hyperparameters and hardware setup}
\subsection{Policy network training parameters}
\begin{itemize}
    \item batch size: 16 goals, 256 premises
    \item number of workers: 8
    \item optimizer: Adam
    \item Adam epsilon: 1e-3
    \item initial learning rate: 1e-4
    \item learning rate decay: exponential, 0.98/100000 steps
    \item embedding size: 128
    \item non-linearity: ReLU
    \item hidden layer dropout: 0.5
    \item GNN hops: 16
    \item layers per hop: 2
    \item initializer range: 0.02
    \item pre-combiner embedding size: 4096
    \item number of combiner layers: 3
    \item ratio of human training data: 0.7 (for human loops), 0.0 (for zero loops)
    \item ratio of historical training data: 0.2 (for human loops), 0.5 (for zero loops)
    \item ratio of fresh training data: 0.1 (for human loops), 0.5 (for zero loops)
\end{itemize}
\subsection{Prover hyperparameters}
Some parameters are picked independently by each prover. These are picked uniformly at random from a given interval, are indicated below as $[n_1, n_2]$. The intervals are inclusive of both end points.
\paragraph{Training provers:}
\begin{itemize}
    \item total number of provers: 2000 (see continuous validation note)
    \item training round interval (each time a model checkpoint is written out): 4000 training steps
    \item prover tree search strategy: BFS
    \item timeout: 300 seconds
    \item maximum number of actions considered (per goal): [10, 30]
    \item maximum \emph{successful} actions (per goal): [10, 18]
    \item maximum number of premises ($k$): [2, 32].
    \item number of premise samples per tactic: 4
    \item tactic timeout: 500 milliseconds
    \item maximum goals explored: 1000000 (practically, no limit)
\end{itemize}

Parameters that vary from experiment-to-experiment:

\begin{tabular}{lllll}
\toprule
\textbf{Loop name}        & \multicolumn{2}{c}{Seed proofs} & \multicolumn{2}{l}{\qquad Premise selection}\\
                          & Human & Generated  & Learnt model & Exploration \\
\midrule
Human reference           & Yes   & -          & Yes          & No ($k_2=0$) \\
Human explore             & Yes   & Exploration & Yes          & Yes ($k_2' \in [0,8]$, $p=0.1$) \\
\midrule
Zero reference            & No    & Reference   & Yes          & No ($k_2=0$)\\
Zero explore              & No    & Exploration & Yes          & Yes ($k_2' \in [0,8]$, $p=0.1$) \\
Zero seeded               & No    & Exploration & Yes          & No ($k_2=0$) \\
Zero hand-engineered only & No    & Exploration & No           & Yes ($k_2=k$, $p=0.1$)\\
Zero explore 0\% dropout & No    & Exploration & Yes          & Yes ($k_2' \in [0,8]$, $p=0.0$) \\
\bottomrule
\end{tabular}

$k_1$ (number of premises to be picked by the network) and $k_2$ (number of premises to be added for exploration) are decided as follows.
In the table above, for experiments where $k_2$ is defined $k_2$ is picked as shown.
For experiments where $k_2'$ is defined, $k_2$ is derived it as $max(\lceil k/2\rceil, k_2')$.
$k_1$ is picked as $k-k_2$.

$p$ is the token dropout probability, defined in Section \ref{sec:retrieval}.

For technical reasons, for the training to start some non-empty training data is needed.
For human RL loops, since human proof data is available, the generated seed data is not strictly necessary.
For zero RL loops, some data needs to be provided.
We generate this data by trying to prove all theorems in the training set on a randomly initialized model (i.e. the 0-th checkpoint of a model).
The hyperparameters used for the provers to generate the seed data is as follows:
\begin{itemize}
        \item prover tree search strategy: BFS
        \item timeout: 1000 seconds
        \item maximum number of actions considered (per goal): 20
        \item maximum \emph{successful} actions (per goal): 5
        \item maximum number of premises: $k=k_1=24, k_2=0$ (reference), $k=k_2=16, k_1=0, p=0.0$ (exploration).
        \item number of premise samples per tactic: 1
        \item tactic timeout: 5000 milliseconds
        \item maximum goals explored: 1000000 (practically, no limit)
\end{itemize}

\paragraph{Validation provers:}
\begin{itemize}
    \item prover tree search strategy: BFS
    \item timeout: 1000 seconds
    \item maximum number of actions considered (per goal): 20 
    \item maximum \emph{successful} actions (per goal): 14
    \item maximum number of premises: 20
    \item number of premise samples per tactic: 4
    \item tactic timeout: 500 milliseconds
    \item maximum goals explored: 1000000 (practically, no limit)
\end{itemize}
Following only apply to continuous validation, not final validation which is run at final checkpoint, and on the full set.
\begin{itemize}
    \item continuous validation interval: 20 rounds (80,000 training steps)
    \item probability to prove a validation theorem: 0.1 (Explanation: the 2000 provers running in the RL loop for training, each independently decides to re-purpose itself with this probability to help with continuous evaluation. With this probability, this leads to over 2000 proof attempts per validation interval on average.)
\end{itemize}

\subsection{Hardware setup}
We used eight NVIDIA Tesla V100 GPUs for distributed training, an additional GPU was used purely for
evaluation, and we maintained a separate parameter server on a CPU machine.
The provers generating training data and running validation run distributed, using 2000 CPUs.

\section{Reinforcement learning loop resources}
\label{apx:loopresources}
Since the improvement of the premise guidance is heavily reliant on generation of data, we run up to 2000 theorem provers distributing the statements each prover is attempting to prove.
Computing predictions takes a few milliseconds but actions in the proof assistant can take up to half a second. We use 8 GPUs for training the policy network and the experience collection uses CPUs only.
Combined with proof search, to have a reasonable chance of proving a statement, we run the theorem prover with a timeout of 5 minutes for
a statement in the training set.
Training over 5 days, a single reinforcement learning loop takes over 25 years of CPU resources and 40 days of GPU resources.

\end{document}